%% file: acl_latex.tex
\pdfoutput=1

\documentclass[11pt]{article}

\usepackage{acl}

\usepackage{times}
\usepackage{latexsym}
\usepackage{algorithm}
\usepackage{algorithmic}
\usepackage{xcolor,colortbl}
\usepackage{booktabs} 

\newcommand{\green}[1]{\colorbox{green!25}{#1}}

\usepackage[T1]{fontenc}

\usepackage[utf8]{inputenc}

\usepackage{microtype}

\usepackage{inconsolata}
\usepackage{multirow}
\usepackage{graphicx}

\title{Arabic Automatic Story Generation with Large Language Models}

\usepackage[colorinlistoftodos,prependcaption,textsize=tiny]{todonotes}

\usepackage{amssymb}
\usepackage{pifont}
\definecolor{blue}{RGB}{0,0,255}
\definecolor{lightblue}{RGB}{0,216,230}

\author{\normalsize Ahmed Oumar El-Shangiti$^{\xi}$~ Fakhraddin Alwajih$^{\lambda}$ \\ ~ \normalsize\textbf{Muhammad Abdul-Mageed}$^{\lambda,\xi,\phi}$~  
 \\
\normalsize $^{\lambda}$ The University of British Columbia\\
\normalsize $^{\xi}$ Mohamed bin Zayed University of Artificial Intelligence (MBZUAI)
\\ 
\normalsize $^{\phi}$ Invertible AI\\
   \texttt{\normalsize ahmed.oumar@mbzuai.ac.ae, muhammad.mageed@ubc.ca}
  }

\begin{document}
\maketitle
\begin{abstract}
\input{sections/abstract}
\end{abstract}

\input{sections/introduction}

\section{Related Work}
\label{section:related work}

\input{sections/related_work}

\section{Data Collection}

\input{sections/data}

\input{sections/prompt_design}

\section{Experiments}
\label{section:Experiments}

\input{sections/experiments}

\section{Evaluation}
\label{section: evaluation}
\input{sections/evaluation}

\section{Conclusion}
\label{section:conclusion}
\input{sections/conclusion}

\section*{Limitations}
\input{sections/limitations}

\section*{Acknowledgments}\label{sec:acknow}
We acknowledge support from Canada Research Chairs (CRC), the Natural Sciences and Engineering Research Council of Canada (NSERC; RGPIN-2018-04267), the Social Sciences and Humanities Research Council of Canada (SSHRC; 435-2018-0576; 895-2020-1004; 895-2021-1008), Canadian Foundation for Innovation (CFI; 37771), Digital Research Alliance of Canada,\footnote{\href{https://alliancecan.ca}{https://alliancecan.ca}} and UBC ARC-Sockeye.
\bibliography{acl_latex}

\appendix

\section{Example Appendix}
\label{sec:appendix}
This is an appendix.
\input{sections/appendix}

\end{document}

%% file: sections/abstract.tex
Large language models (LLMs) have recently emerged as a powerful tool for a wide range of language generation tasks. Nevertheless, this progress has been slower in Arabic. In this work, we focus on the task of generating stories from LLMs. For our training, we use stories acquired through machine translation (MT) as well as GPT-4. For the MT data, we develop a careful pipeline that ensures we acquire high-quality stories. For our GPT-4\footnote{GPT-4 refers to GPT-4-0125-preview} data, we introduce crafted prompts that allow us to generate data well-suited to the Arabic context in both Modern Standard Arabic (MSA) and two Arabic dialects (Egyptian and Moroccan). For example, we generate stories tailored to various Arab countries on a wide host of topics. Our manual evaluation shows that our model fine-tuned on these training datasets can generate coherent stories that adhere to our instructions. We also conduct an extensive automatic and human evaluation comparing our models against state-of-the-art proprietary and open-source models. Our datasets and models will be made publicly available at \url{https://github.com/UBC-NLP/arastories}.

%% file: sections/introduction.tex
\section{Introduction}
Storytelling is an essential human skill that serves to transmit knowledge, impart values, and connect individuals through tales of daily experiences. It is utilized in education, where teachers harness children's natural affinity for stories to foster cognitive and literacy development. Additionally, stories and legends, viewed as cultural heritage, are passed down through generations by parents, enriching the culture and preserving traditions.

The role of storytelling also extends beyond its traditional roots; it acts as a vital connection between the primitive oral language skills in early childhood and the advanced language abilities associated with literacy. As such, the task of automatic story generation presents numerous benefits across different fields. In entertainment, it allows for the efficient creation of diverse narratives~\cite{xie2024creating}. In education, tailored stories can be crafted to address the unique needs of learners. In gaming, interactive storytelling significantly enhances user engagement and enjoyment~\cite{patel2024swag}. And these are only a few application domains.

Progress in natural language processing (NLP) technologies, particularly with large language models (LLMs) such as GPT-4 and Gemini, has made automatic story generation both viable and effective, producing stories with notable fluency and coherence. While substantial efforts have been made to advance automatic story generation in English using generative models, the development of such technologies for Arabic has been limited due to a scarcity of Arabic short story data and minimal focus from the research community.

In this study, we present a novel approach to automatic story generation utilizing the powerful Arabic LLM, AraLLaMA~\cite{alwajih2024peacock}. We enhance AraLLaMA through fine-tuning with both translated and synthetic datasets to optimize its story-generating capabilities. We explore two fine-tuning strategies: one involving direct application of a synthetic dataset produced by GPT-4, and another beginning with an analogous synthetic dataset translated from English. Additionally, we extended the model's utility by fine-tuning it with data from two Arabic dialects, enabling the generation of stories in both Modern Standard Arabic (MSA) and these two dialects. The efficacy of our model is assessed through human evaluation, which confirmed its ability to produce coherent and fluent narratives as per specified instructions.

Our contributions are manifold, summarized as follows:

\begin{enumerate}
    \item We introduce powerful models capable of generating coherent and fluent stories in MSA and two Arabic dialects.
    \item We offer a newly created framework for Arabic automatic story evaluation based on LLMs.
    \item We develop two novel datasets for automatic story generation: one consisting of translated narratives from the TinyStories~\cite{eldan2023tinystories} dataset, which was meticulously curated, and another comprising a synthetic dataset created using GPT-4, featuring narratives in MSA and two dialects.
    \item We compare two distinct fine-tuning methods on AraLLaMA against AceGPT-7B~\cite{huang2024acegpt}, GPT-3.5, and Command-R\footnote{\url{https://dashboard.cohere.com/playground/chat?}}, powerful open source and proprietary models using extensive automatic and human evaluations.
\end{enumerate}
The remainder of this paper is organized as follows: Section \ref{section:related work} provides a review of prior studies focusing on the task of automatic story generation. Section \ref{section:data collection} details the creation of our datasets. In Section \ref{section:Prompt Design}, we outline our prompt design. In section \ref{section:Experiments} we detail our different experiments. Results and key insights from our comparative analysis of our fine-tuned models against various commercial and open-source models are discussed in Section \ref{section: evaluation}. The paper concludes with Section \ref{section:conclusion}.

%% file: sections/related_work.tex
\subsection{Early Work on Story Generation.} \citet{jain2017story} is an early work on generating coherent stories, experimenting with two paradigms: Statistical Machine Translation (SMT) and Deep Learning. SMT treats story generation as a translation task, while Deep Learning uses Recurrent Neural Networks (RNNs) to encode sequences of input descriptions into hidden representations, which are then transformed into detailed summaries. They evaluate their models using BLEU, ROUGE-L, and human evaluation. \citet{fan-etal-2018-hierarchical} propose a hierarchical model that first generates a story premise using a convolutional language model \cite{dauphin2017language} and then a seq2seq model to create a story that follows the premise. They incorporate gated multi-scale attention and model fusion to improve prompt adherence.

\citet{akoury2020storium} introduce the STORIUM dataset\footnote{\url{https://storium.com/}} and fine-tune GPT-2-medium \cite{radford2019language} for generating short story scene entries, motivated by GPT-2's 1024-token context window. Plug-and-Blend \cite{lin2021plugandblend} consists of a Blending Generative Model (BGM) and Planner for controllable story generation. BGM facilitates controlled continuations, while Planner specifies control parameters based on topic descriptions and story sections. The authors fine-tune GPT-2-large \cite{radford2019language} on ROCStories \cite{mostafazadeh-etal-2016-corpus} and use pre-trained GeDi \cite{krause2020gedi} as the guiding model, evaluating fluency and fidelity through human evaluation.

\subsection{LLM Story Generation.}

\citet{mirowski2022cowriting} propose using a 70B Chinchilla LLM called Dramatron for generating long narratives, such as full scripts and screenplays, through prompting, prompt chaining, and hierarchical generation. Dramatron supports collaborative writing and was qualitatively assessed via co-writing sessions and interviews with 15 industry professionals.

\citet{yang-etal-2022-re3} propose the Recursive Reprompting and Revision (Re3) framework automatically generates longer stories without human intervention, distinguishing it from previous approaches. Re3 comprises four modules: Plan, Draft, Rewrite, and Edit. The Plan module creates a story plan using GPT-3 \cite{brown2020language} to add details to a given premise. The Draft module generates story continuations by recursively prompting GPT-3, dynamically updating the prompt with information from the plan and story. The Rewrite module reranks alternate continuations to select the best ones, and the Edit module ensures factual consistency with earlier parts of the story. Re3 operates in a zero-shot manner, allowing it to generate longer stories without domain constraints.

\citet{patel2024swag} propose a creative storytelling framework with two components: the story generation model and the Action Discriminator model (AD LLM). They train these models in a feedback loop called SWAG. Initially, a prompt is used to generate the first paragraph, which is then fed into the AD LLM with actions (e.g., "add suspense") to produce the best continuation. This process is repeated until the story reaches the desired length. The model is trained using Direct Preference Optimization (DPO) \cite{rafailov2023direct}, with preference data generated by GPT-4~\cite{openai2024gpt4} and Mixtral-8×7B \cite{jiang2024mixtral}. GPT-4 samples are chosen, while Mixtral-8×7B samples are rejected. Evaluation is conducted using both human and GPT-4 assessments.

\citet{xie2024creating} introduces a method for generating suspenseful stories with LLMs using iterative prompting based on psychological and narratological theories of suspense. This zero-shot approach does not require pre-existing story corpora. Human evaluations demonstrate the effectiveness of this technique in crafting engaging suspenseful stories, and controlled studies explore factors influencing readers' perception of suspense.

\citet{radwan2024sard} introduces SARD, a tool with a visual drag-and-drop interface for creating multi-chapter stories using advanced large language models. Wordcraft~\cite{10.1145/3490099.3511105} is a web application for story writing that combines a text editor with controls for prompting an LLM to perform various story-generation tasks.

\subsection{Evaluation of Story Generation in Literature}
There are basically two types of evaluations for story generation in the literature: human evaluation and automatic evaluation. We explore these evaluation methods in the following subsections:
\subsubsection{Human Evaluation}
~\citet{akoury2020storium} integrate their fine-tuned model into the STORIUM collaborative storytelling platform, where real authors can query the model to generate suggested story continuations. The authors could edit the generated text by adding or deleting content. The edited stories were collected along with ratings from the authors' on properties such as relevance, fluency, coherence, and likability. They also propose a new automatic metric called User Story Edit Ratings (USER), inspired by the longest common subsequence (LCS) of the ROUGE metric~\cite{lin-2004-rouge}, which measures how much of the generated text is preserved in the edited version. 

The authors of Re3~\cite{yang-etal-2022-re3} ask workers from Amazon Mechanical Turk to rate Re3-generated stories against GPT-3 and GPT-3 fine-tune on stories from the WritingPrompts dataset. The evaluation criteria include interestingness, coherence, fluency, human-likeness, and relevance. Workers also identify shortcomings in the generated stories, such as disfluency, repetitiveness, confusing inconsistencies, and narration problems. Re3 outperforms all baselines on almost all criteria.

~\citet{xie2024creating} rely on a pool of three human studies to evaluate their framework for suspenseful story generation. In the first study, human judges compare stories generated by their approach against those generated by a strong baseline (ChatGPT) based on suspense, novelty, enjoyment, logical sense, and naturalness. The second study involved ablations on their system compared against the full system. In the third phase, participants reviewed the story's structure to verify internal processes. Ninety participants assessed 30 story pairs, with each pair reviewed by 30 participants. Their approach outperforms all baselines on all criteria except for a 56\% tie with ChatGPT on naturalness.

The authors of~\cite{patel2024swag} evaluate their story generation method using human judges and GPT-4. Surge AI employees assessed 50 stories generated by the Llama-2-7B and Mistral-7B models, enhanced by the SWAG technique, against four baselines: the end-to-end approach, a random selection method, GPT-3.5-turbo, and GPT-4-turbo. Evaluations focused on interestingness, surprise, and coherence. The findings show a preference for SWAG-enhanced stories over conventional methods by both human judges and GPT-4, with SWAG models winning 61.5

~\citet{wang2024weaver} compare Weaver's variations against other open-source and proprietary LMs, including GPT-4, GLM-4, ERNIE-Bot-4.0, and Gemini-pro. Evaluations by human professionals and GPT-4 were based on creativity, style, relevance, and fluency. Weaver-Ultra was preferred 1576 and 1657 times out of 3540 samples by humans and GPT-4, respectively.


\subsubsection{Automatic Evaluation}
Evaluating creative writing such as story generation is a challenging task.~\citet{jain2017story}, one of the earlier works on neural-based story generation, uses machine translation metrics (BLEU-4, METEOR, TER, and ROUGE-L) to evaluate story generation. The overall results were low, with SMT-based methods scoring better on BLEU-4 than seq2seq models, despite being less coherent. The scores were 3.5 and 1.98 for SMT and seq2seq models, respectively, indicating that n-gram-based metrics are not suitable for creative writing judgment. GPT-Eval, an evaluation framework based on GPT-4 \cite{eldan2023tinystories}, takes in a story and provides a general assessment and a score out of 10 in four criteria: grammar, creativity, consistency, and age group.

\subsection{Common Datasets for Story Generation}
\input{tables/datasets}
To provide an overview of the resources used in story generation research, we summarize the most common datasets used to build automatic systems for generating stories in Table~\ref{tab:datasets_desc}. These datasets vary in size, nature, availability, and average length of the stories they contain. The datasets include human-generated stories as well as those created by advanced language models like GPT-3.5 and GPT-4. They are valuable resources for training and evaluating story-generation models.

\subsection{Arabic Story Generation}
The task of automatic story generation is uninvestigated in the Arabic NLP community. However,  ~\cite{alhussain2024crosslingual} utilize cross-lingual transfer learning to address the scarcity of Arabic data in Story Ending Generation (SEG) task by leveraging English story corpora. 

%% file: tables/datasets.tex
\begin{table*}[ht!]
\centering
\begin{tabular}{lllll}
\hline
\textbf{Dataset name} & \textbf{Size} & \textbf{Nature} & \textbf{Available} & \textbf{Avg Length}\\
\hline
\citet{huang-etal-2016-visual} & $41K$ & Human & No & $21.2$ Tokens\\

ROCStories & $50K$ & Human & Yes & 5 sentences\\
WritingPrompts & $300K$ & Human & Yes &$ 734.5$ Words\\
STORIUM & $6K$ & Human & Yes &  $19K$ Tokens\\
TinyStories & $5M$ & GPT-4/GPT-3.5 & Yes  \\
Weaver~\cite{wang2024weaver} & $500K$ & GTP-4/GPT-3.5 & No & ---\\
SWAG~\cite{patel2024swag} & $20K$ & GPT-4/Llama-2-7B/Mistral-2-7B & No & $5K$ Words\\
\hline
\end{tabular}
\caption{Description of the stories datasets}
\label{tab:datasets_desc}
\end{table*}

%% file: sections/data.tex
\label{section:data collection}


We compile data from different resources. We first translate $1.13$M English stories generated by GPT-4 alongside their prompts from the \textit{TinyStories} dataset~\cite{eldan2023tinystories} using Google translate API.\footnote{\url{translate.googleapis.com}} To ensure that we have only high-quality translation, we apply a filtering strategy based on multilingual sentence embeddings~\cite{feng-etal-2022-language} and remove the story pairs whose embedding similarity is less than $92$\%. 

\subsection{Filtering Strategy}
With the aim to train only on high-quality data, we apply Algorithm~\ref{alg:filtering_stories} to our dataset. The final threshold was $92\%$. And we were able to maintain $545$K samples which represents $48.3\%$ of the translated data.
\begin{algorithm}
    \caption{Filtering Stories Based on Similarity Score}\label{alg:filtering_stories}
    \begin{algorithmic}[1]
        \REQUIRE Stories dataset \( D \), Similarity threshold \( t \), Minimum word count \( m = 50 \)
        \STATE Remove stories shorter than \( m \) words
        \STATE Sort stories based on similarity score
        \STATE Filter out stories whose similarity with the original story is less than the threshold \( t \)
        \STATE Get a human in the loop to manually check some random samples
        \STATE Set a new threshold \( t' \)
        \STATE Repeat steps 2-5 until satisfactory translation quality is achieved
    \end{algorithmic}
\end{algorithm}

\subsection{Generated Data}

We also generate our own stories from GPT-4-Turbo API using a carefully designed set of prompts and features (see Section \ref{section:Prompt Design}). The dataset generated with our prompt template is in three Arabic varieties, namely MSA, Moroccan, and Egyptian. We tested the ability of GPT-4-Turbo to generate other Arabic dialects, but the generated content was mostly MSA. For this reason, we decided to limit our work to the above-mentioned varieties. We generate $1,000$ stories for each variety, making a total of $3,000$ stories. We also create $20$ additional prompts for evaluation. We provide an example of stories generated by GPT-4-Turbo using our custom prompt template in Figure~\ref{apdx:training_examples}.

%% file: sections/prompt_design.tex
\section{Prompt Design}
\label{section:Prompt Design}
Prompting is an approach employed by users to interface with LLMs~\cite{white2023prompt}. It functions as the primary mode of communication with these models, effectively serving as the input language that LLMs are designed to interpret and respond to. The efficacy of the generated output is significantly correlated with the quality and structure of the input prompt. This relationship underscores the critical role that prompt engineering plays in optimizing LLM performance and output relevance.
In our context, the prompt can be conceptualized as a set of instructions or parameters that guide the LLM's for the Arabic story generation process. The prompt's composition, including its specificity, clarity, features, and relevance to the desired output, directly influences the model's ability to generate appropriate and accurate stories that adhere to our instructions. This causal relationship between prompt quality and output quality highlights the importance of developing sophisticated prompting strategies to fully leverage the capabilities of LLMs for Arabic story generation.
\\
\subsection{Initial Investigation}
When we started this study, we had three prompting choices. We either prompt in English, Arabic, or dialect for dialectal stories. Recent studies have shown that LLMs perform better when prompted in English compared to other languages~\cite{etxaniz2023multilingual,kadaoui2023tarjamat}. However, we still wanted to test these claims and verify if they hold in our case. For this reason, we query GPT-4 with prompts in English and Arabic respectively and compared the corresponding generated stories. In this exploratory stage, we manually looked into each story and checked the creativity, fluency, and instruction following of the model. This pilot study did not reveal any critical differences between both languages in terms of fluency and creativity of the Arabic-generated stories. 

For further investigation, we carry out the same experiment comparing MSA prompts versus dialectal prompts for dialectal story generation. This time, we ran $10$ samples to AlDi~\cite{keleg-etal-2023-aldi} ($5$ generated with an MSA prompts while the other $5$ are generated with dialectal prompts) to quantify the level of dialectness of each generated story. The scores were 81.84\% and	81.6\%	for the percentage of dialectness of the content generated with MSA and dialectal prompts respectively. Given that the difference are not significant, we decided to create prompts in each Arabic variety. This choice is mainly motivated by our intent to make the story and the prompt uniform as well as making it easier for the user to write their prompt directly in the chosen variety without a need to be fluent in MSA. Next, we describe the details of our prompt template.

\subsection{Prompt Template} We design our prompt template with two goals in mind. These are (i) to ensure high quality of the generated output and (ii) make the generated output as diverse as possible. 
To ensure the variety of the generated stories, we carefully design a set of 12 features: \textit{\{age, place, end of story, dialogue, number of characters, moral of the story, topic, country, season, activity, emotion, plot twist\}}. Our template is designed in such a way that each feature has a probability $p$ of appearance in a particular prompt. Meaning some features might be present in a prompt while others are not. Except for the following features where they appear in each prompt: \textit{age, number of characters}, and \textit{country}. Based on our preliminary observations, GPT-4 is able to generate coherent stories from dialectal prompts (i.e., Egyptian and Moroccan). Hence, we ask two native speakers to translate our prompt template, originally written in MSA to Egyptian and Moroccan dialects. This prompt template is used to generate MSA and dialectal stories from GPT-4 and later to fine-tune our models.

%% file: sections/experiments.tex
We conduct two sets of experiments:
\begin{enumerate}
\item Directly fine-tuning on the generated data from GPT-4 Turbo-preview.
\item Fine-tuning on translated data, followed by further fine-tuning on data generated with the GPT-4-Turbo-preview model.
\end{enumerate}

The details of each experiment are described next.


\subsection{Supervised Fine-Tuning (SFT)}
\label{exp:1}
\begin{figure*}[ht!]
    \centering
    \includegraphics[width=\linewidth]{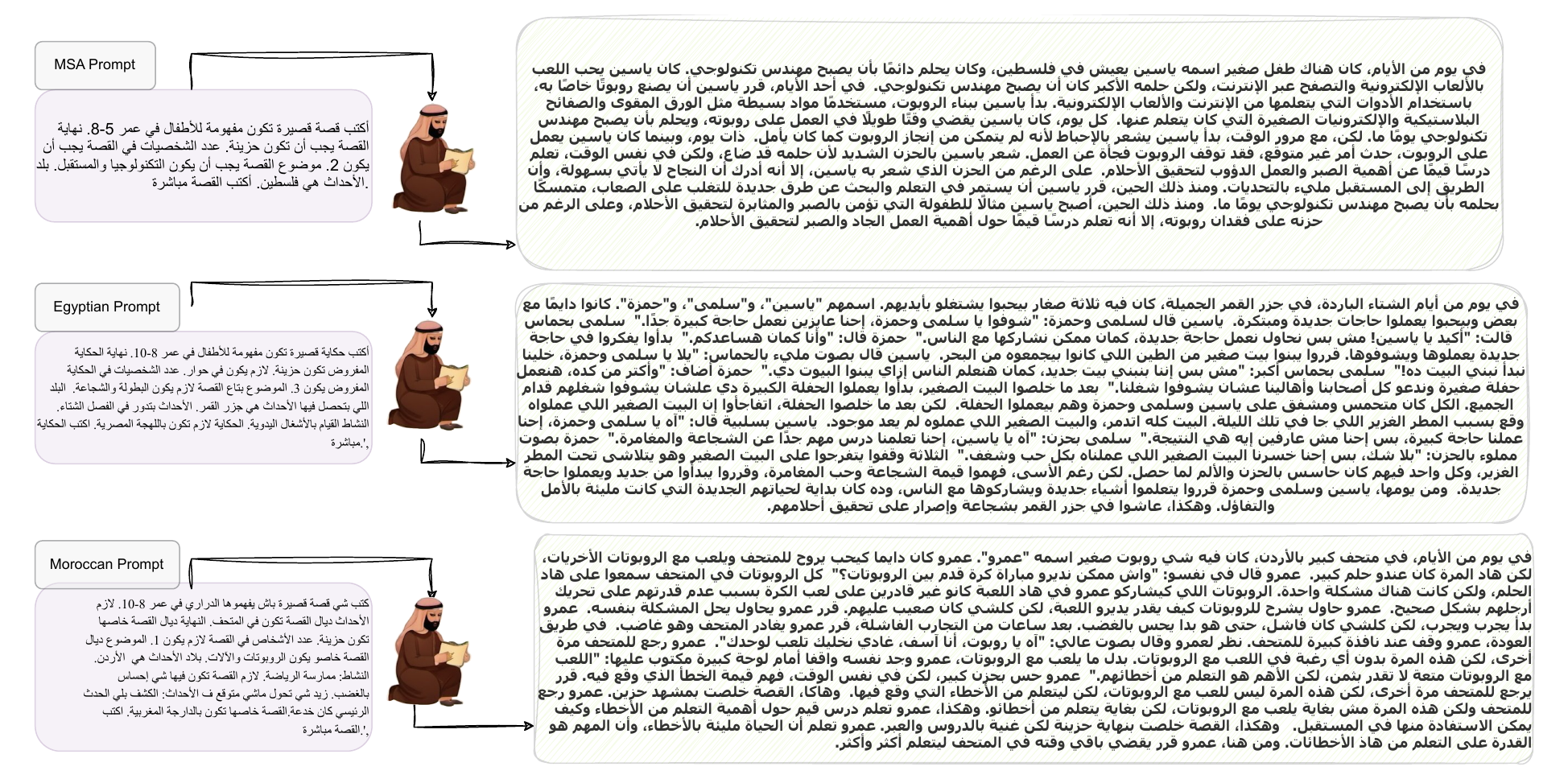}  
    \caption{Samples of different stories generated with our models for the three Arabic Varieties.}
    \label{fig:examples of data generated with our models.}
\end{figure*}
We instruct fine-tune AraLLaMa-2-base~\cite{alwajih2024peacock} using a diverse Arabic instruction tuning dataset generated with our custom prompt template. AraLLaMa-2-base is a 7B parameter model based on Llama-2~\cite{touvron2023llama}, continually pre-trained on Arabic data.  AraLLaMa-2 has shown superior performance compared to other Arabic LLMs such as AceGPT-7B~\cite{huang2024acegpt} and Jais-7B~\cite{sengupta2023jais}, hence we  adopt it for our experiments. For computational efficiency, all our models are trained with Huggingface PEFT library~\cite{peft}. In each experiment, our base model, AraLLaMa-2 is quantized in 4-bit precision and then a new QLoRA layer~\cite{dettmers2023qlora} is added. During our experiments, we keep the base model frozen and update the QLoRA layer only. The instruction fine-tuning is performed by updating a newly added QLoRA layer, with $\alpha$ set to $16$, $r$ set to $64$, QLoRA layer dimension set to $64$, gradient accumulation at $10$, batch size equal $1$. We use as optimizer AdamW~\cite{loshchilov2019decoupled}, a dropout is equal to $10\%$, a learning rate set to $4*10{e-5}$. We train for $20$ epochs. This training took approximately $5.5$ hours on a single Nvidia A100 GPU. We call the model artifact resulting from this training \textbf{Model A}.

\subsection{Two-Step Fine-Tuning}

This experiment is divided into two steps. First, we instruct fine-tune AraLLaMa-2-base on large-scale translated data from  the \textit{TinyStories} dataset~\cite{eldan2023tinystories} for $15,000$ steps. The second step is taking the trained model from the previous step and further instruct fine-tuning it on a smaller dataset from experiment 1 (Section \ref{exp:1}). The hyperparameters are the same as in the previous experiment. The overall training took about $17.5$ hours on the Nvidia A100 GPU. We provide examples of stories generated with our models across all three Arabic varieties in Figure~\ref{fig:examples of data generated with our models.}. We call the model artifact resulting from this training \textbf{Model B}.

%% file: sections/evaluation.tex
\input{tables/evaluation_msa}

\begin{figure}[h]
\centering
\includegraphics[width=1\linewidth]{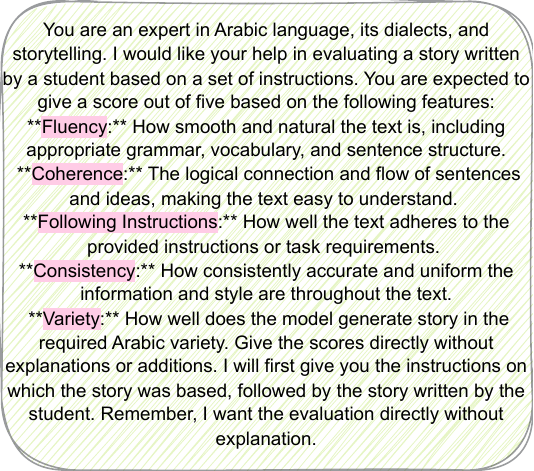}
\caption{The prompt we pass to GPT-4-Turbo to evaluate stories generated with different models.}
\label{fig:GPT-4-evaluatin-prompt}
\end{figure}

Evaluating generative tasks remains an open problem in the AI community. However, in our study, we follow previous works such as~\cite{eldan2023tinystories} in adapting GPT-4 as an evaluator for model performance. We also conduct an extensive human evaluation trying to understand different model capabilities and how the evaluation of GPT-4 compares to that of human judges. We next describe our two evaluation strategies.

\subsection{GPT-4 As a Judge}
\label{subsection:human evaluation}
\begin{figure*}[ht!]
    \centering
    \includegraphics[width=\linewidth]{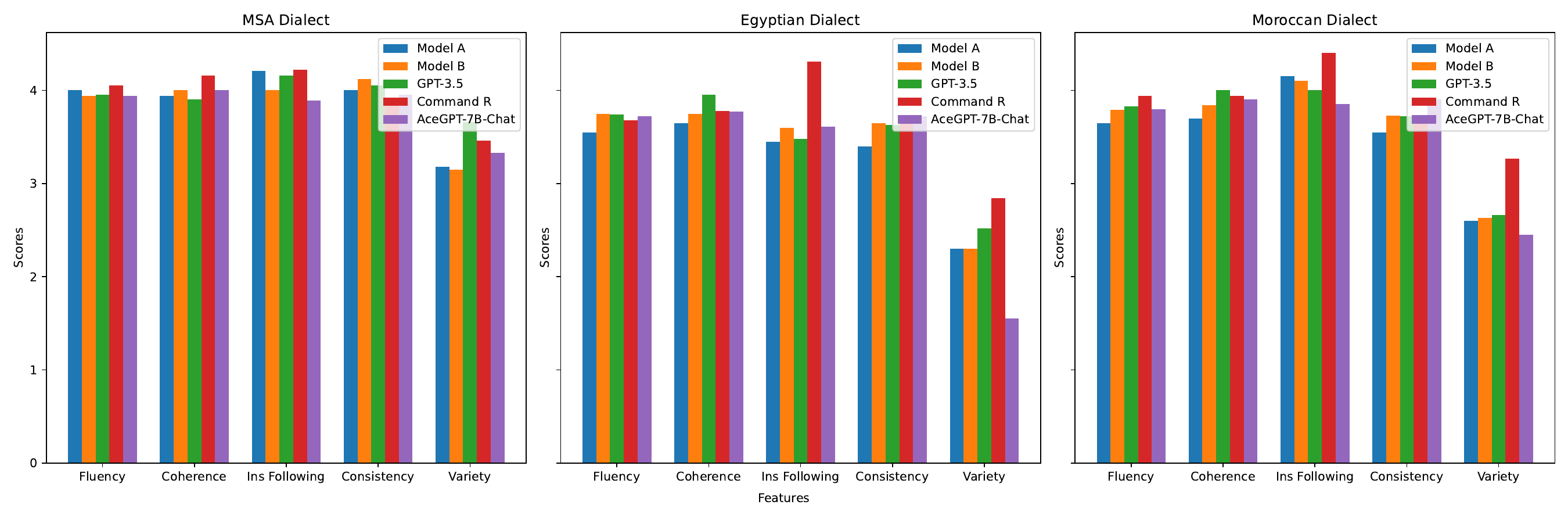}
    \caption{Models performance across the MSA, Egyptian, and Moroccan varieties.}
    \label{fig:model_performance_bar_chart}
\end{figure*}


We evaluate our models on five criteria scored by GPT-4. We design a comprehensive prompt that works as follows: given the original story prompt plus the corresponding generated story, we ask GPT-4 to act as an Arabic language expert and  assign a score out of five on the following criteria:
\begin{itemize}
  \item \textbf{Fluency:} The degree to which the text reads smoothly and naturally, with appropriate grammar, vocabulary, and sentence structure.
    \item \textbf{Coherence:} The logical connection and flow between sentences and ideas, making the text easy to understand.
    \item \textbf{Instruction Following:} The extent to which the text meets the given instructions or task requirements.
    \item \textbf{Consistency:} The degree to which the information and style within the text remain uniform and accurate throughout.
    \item \textbf{Variety:} How well the model generates a story in the
required Arabic variety.   
\end{itemize}
We find that GPT-4 tends to score the MSA content higher than dialectal ones, even if the task is to generate a dialectal story. To mitigate this issue, we added this last feature where we explicitly ask GPT-4 if the generated content follows the required Arabic variety specified in the prompt or not, and to which degree. Figure~\ref{fig:GPT-4-evaluatin-prompt} demonstrates the prompt we pass to GPT-4 for evaluation.\\
We evaluate 20 new prompts by asking the models to generate 20 corresponding stories and then pass the prompt plus story to GPT-4 for evaluation. We compare our two models against three other open and proprietary models. Namely, we compare against GPT-3.5, Command-R\footnote{\url{https://dashboard.cohere.com/playground/chat?}}, and AceGPT-7B-Chat~\cite{huang2024acegpt}.

It is pertinent to note that we experimented with other strong open-source models, such as LLaMA-3-70B-Chat~\cite{touvron2023llama} and Mixtral-8x7B~\cite{jiang2024mixtral} (accessed through an API), but these models failed to adhere to our instructions. This failure highlights the superiority of our models over these strong baselines. Furthermore, we could not compare our models against larger Arabic LLMs, such as Jais-30B~\cite{jain2017story} and AceGPT-13B~\cite{huang2024acegpt}, due to computational constraints. In other words, we limited our comparisons to 7B models unless an API was available.

Table \ref{tab:GPT-4-evaluation-results} depicts the results of each tested model according to GPT-4. As we clearly see from Table \ref{tab:GPT-4-evaluation-results}, the overall results gap between the models is very narrow. In addition,  
 both our model A and model B are very competitive with larger models even though they are an order of magnitude smaller. Model A performs well in \textit{Instruction Following} across all three varieties and shows strong \textit{Consistency} and \textit{Fluency} in MSA. Model B exhibits better Consistency in MSA and Moroccan, shows strong Fluency and \textit{Coherence} in Egyptian, and relatively lower \textit{Variety} scores.
 
 Model B which 
was exposed to additional training steps on translated data, performs better than Model A on \textit{almost all metrics across dialects}, which proves indeed the intuition behind training on more data does help. Comparative results suggest that there might be opportunities for further fine-tuning or learning from stronger models such as Command-R since this model shows strong performance across multiple metrics. Our results demonstrate that Command-R is the strongest baseline, and the most consistent model across metrics and varieties. Even in the \textit{Variety} feature where the performance of other models falls, Command-R achieves a score as high as $3.27$. Command-R outperforms even GPT-3.5 while being only a fraction of its size, suggesting that the model size is not everything and the quality of data does help. We can see from Table \ref{tab:GPT-4-evaluation-results} that almost all metrics drop for dialectal varieties compared to MSA. This can be directly linked to the lack of Arabic dialectal data that LLMs have been exposed to during the pre-training stage. We included bar charts in Figure ~\ref{fig:model_performance_bar_chart} for more details.
 
\subsection{Human Evaluation}
\begin{figure}[ht!]
    \centering
    \includegraphics[width=\columnwidth]{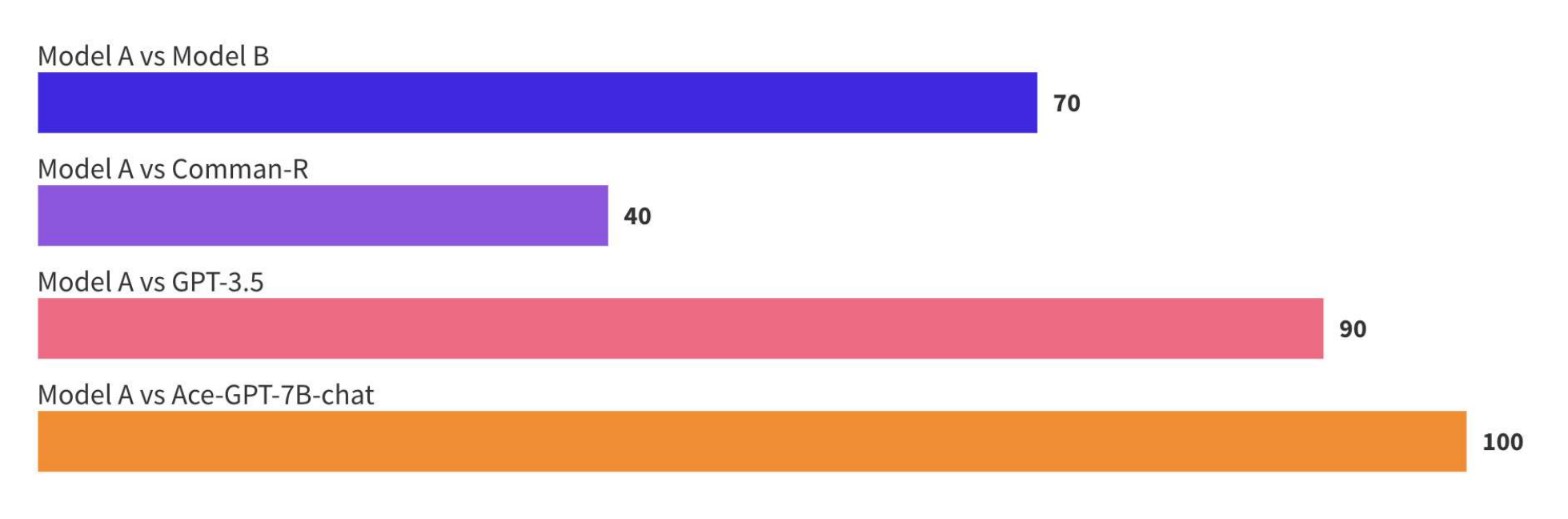}
    \caption{\textbf{Model A} vs. other models in MSA human evaluation. Numbers reflect the number of times Model A is preferred over other models.}
    \label{fig:our_model_vs_different_models_msa}
\end{figure}

\begin{figure}[ht!]
    \centering
    \includegraphics[width=\columnwidth]{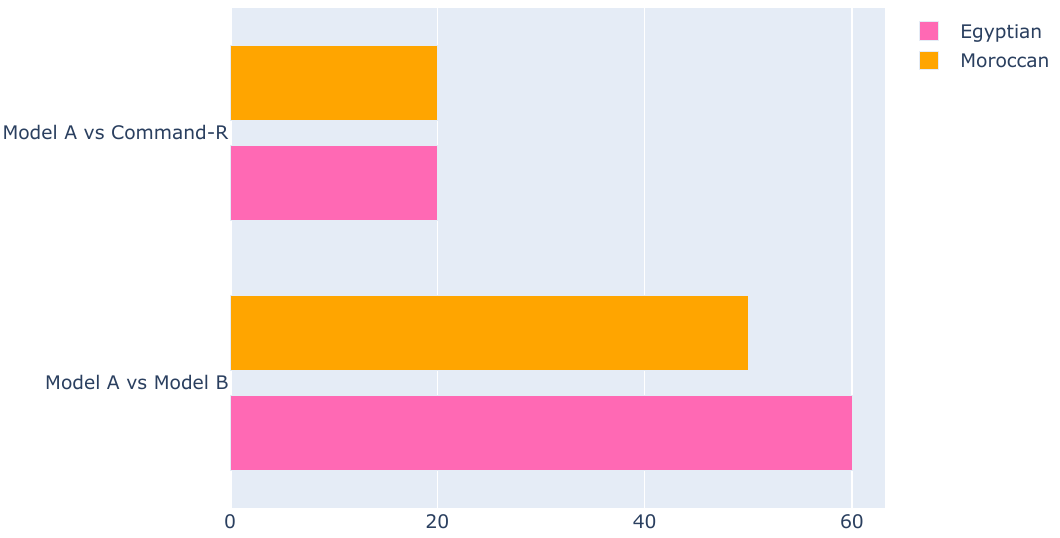}
    \caption{Human comparison of Model A vs Model B and Command-R on Moroccan and Egyptian dialectal story generation. the numbers reflect the number of times Model
A was preferred over other models in percentage. }
    \label{fig:human_eval_egy_mor_vis}
\end{figure}

\noindent \textbf{Pilot evaluation.} We carry out a pilot investigation where a native speaker of Arabic with knowledge of different dialects inspects stories generated by different models and comes up with observations. The expert finds that Model A tends to generate longer stories compared to Model B. Both models A and B are less perfomant in the Moroccan dialect compared to the Egyptian dialect. AceGPT-7B-Chat~\cite{huang2024acegpt} and GPT-3.5 fail to generate dialectal stories, even when explicitly prompted to do so.
\\
{\bf Extensive human comparison of different models.} We ask four Arabic native speakers to rank ten stories generated by different models based on the following criteria: \textit{Instruction Following}, \textit{Fluency}, and \textit{Variety Adherence}. However, since AceGPT-7B-Chat and GPT-3.5 failed to generate dialectal content, we include these models only in the MSA part of the human evaluation task (i.e., we exclude them from the dialectal evaluation). We thus compare our two models, model A and model B, to Command-R. We ask an Arabic native speaker to compare different models to each other, finding our Model A to surprisingly be almost always better than GPT-3.5($90\%$). Our model A also outperforms Command-R(35B) $40\%$ of times, always outperforms AceGPT-7B-Chat, and Model B $70\%$. More details are in Figure ~\ref{fig:our_model_vs_different_models_msa}.

For the human evaluation of dialects, we compare our models against Command-R, which was the only model other than ours able to generate dialectal content. 
Our experts find that Model A to be able to outperform Command-R $20\%$ of the time for both dialects and outperforming Model B $50\%$ and $60\%$ on Moroccan and Egyptian dialectal stories, respectively (Figure ~\ref{fig:human_eval_egy_mor_vis} for visualization).


%% file: tables/evaluation_msa.tex
\begin{table*}[ht!]
    \centering
    \resizebox{\textwidth}{!}{
    \begin{tabular}{llcccccc}
        \toprule
        \textbf{Dialect} & \textbf{Model} & \textbf{Model Size} & \textbf{Fluency} & \textbf{Coherence} & \textbf{Ins Following} & \textbf{Consistency} & \textbf{Variety}\\
        \midrule
        \multirow{5}{*}{MSA}
        & Model A & 7B & 4.0 & 3.94 & 4.21 & 4.0 & 3.18\\
        & Model B & 7B & 3.94 & 4.0 & 4.0 & \green{4.12} & 3.15 \\
        & GPT-3.5 & Unk & 3.95 & 3.9 & 4.16 & 4.05 & \green{3.66} \\
        & Command-R & 35B & \green{4.05} & \green{4.16} & \green{4.22} & 3.88 & 3.46 \\
        & AceGPT-Chat & 7B & 3.94 & 4.00 & 3.89 & 3.95 & 3.33\\
        \midrule
        \multirow{5}{*}{Egyptian} 
        & Model A & 7B & 3.55 & 3.65 & 3.45 & 3.40 & 2.30 \\
        & Model B & 7B & \green{3.75} & 3.75 & 3.60 & 3.65 & 2.30 \\
        & GPT-3.5 & Unk & 3.74 & \green{3.95} & 3.48 & 3.63 & 2.52 \\
        & Command-R & 35B & 3.68 & 3.78 & \green{4.31} & \green{3.73} & \green{2.84} \\
        & AceGPT-Chat & 7B & 3.72 & 3.77 & 3.61 & 3.72 & 1.55\\
        \midrule
        \multirow{5}{*}{Moroccan} 
        & Model A & 7B & 3.65 & 3.70 & 4.15 & 3.55 & 2.60 \\
        & Model B & 7B & 3.79 & 3.84 & 4.10 & \green{3.73} & 2.63 \\
        & GPT-3.5 & Unk & 3.83 & \green{4.0} & 4.0 & 3.72 & 2.66 \\
        & Command-R & 35B & \green{3.94} & 3.94 & \green{4.4} & 3.72 & \green{3.27} \\
        & AceGPT-Chat & 7B & 3.8 & 3.9 & 3.85 & 3.9 & 2.45\\
        \bottomrule
    \end{tabular}
    }
    \caption{Results of our Two Models Across three Arabic varieties scored by GPT-4. \textbf{Model A} is AraLLaMa-base instruction fine-tuned on data generated from GPT-4-Turbo. \textbf{Model B} is AraLLaMa-base fine-tuned on the translated data then on the data generated from GPT-4-Turbo. The scores are on a scale of five points. The best scores are highlighted in green. The size denotes the model size in billions, GPT-3.5 size is unknown.}
    \label{tab:GPT-4-evaluation-results}
\end{table*}

%% file: sections/conclusion.tex
In this paper, we present the first LLM-based study on automatic Arabic story generation. Our study takes as its target MSA and two Arabic dialects (Egyptian and Moroccan). For our purpose, we translate and generate datasets based on a custom prompt template. We fine-tune our models on these datasets, comparing against both equally- and bigger-sized models. Through extensive automatic and human evaluation, we empirically show our models' superiority to  strong baselines. In the future, we plan to fine-tune bigger models on larger datasets. We also plan to include more dialects in our training, for wider coverage.

%% file: sections/limitations.tex
This study has the following limitations:

\begin{itemize}
    \item \textbf{Compute constraints.} Due to computational limitations, we restricted ourselves to models with a maximum size of seven billion parameters or those with an available API.
    \item \textbf{Limited data.} Our training dataset consisted of only 3,000 samples of high-quality data generated by GPT-4. In the future, we are planning to generate more data with the newly released GPT-4o.
    \item \textbf{Lack of error analysis:} We believe carrying out an error analysis would benefit our work. In particular, we observe that GPT-4 does not fully adhere to our instructions 100\% of the time during the generation of training data. This could lead to issues in the data generated using this model and an error analysis could uncover any such limitations. Future work should take this into acount.
\end{itemize}

\section*{Ethical Considerations}
Similar to other generative models, our model can reflect the bias in its data. Any use of the model should take this into account. 

%% file: sections/appendix.tex
We provide an addition organized as follows:

\begin{itemize}
\item Training Examples \ref{apdx:training_examples}

\end{itemize}

\subsection{Training Examples}
\label{apdx:training_examples}

\begin{figure*}[ht]
    \centering
    \includegraphics[width=\linewidth]{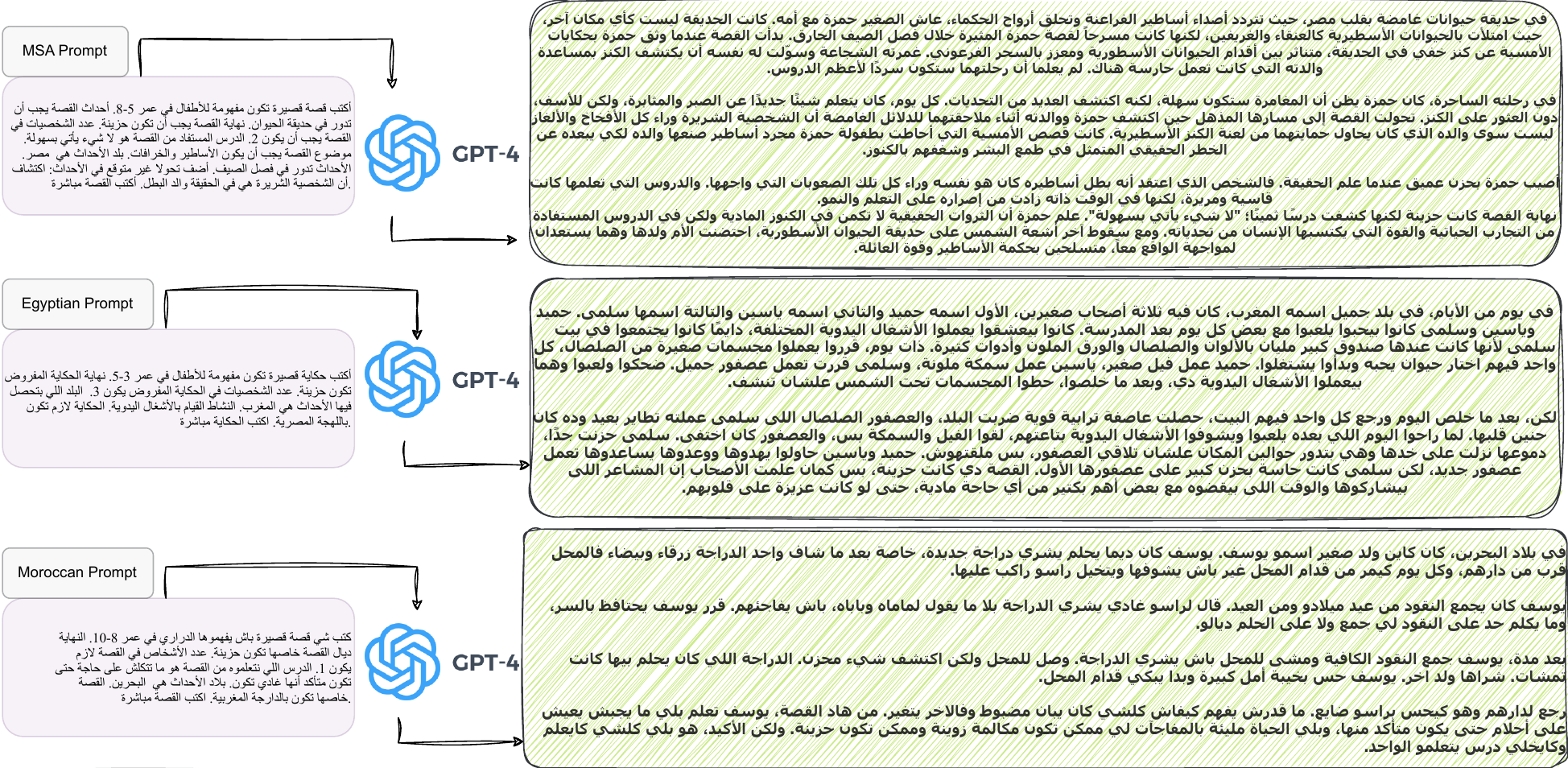}  
    \caption{Example of our training samples generated with GPT-4-Turbo. The figure depicts prompts and their corresponding stories in three Arabic varieties: MSA, Egyptian, and Moroccan dialects, correspondingly.}
    \label{fig:example of generated data with gpt4}
\end{figure*}